\documentclass{article}

\usepackage{iclr2024_conference,times}

\usepackage{fullpage}

\usepackage{algorithm}
\usepackage{algorithmic}
\usepackage{amsthm}
\usepackage{amsmath}
\usepackage{amssymb}
\usepackage{mathtools}
\usepackage{graphicx}
\usepackage{subfigure}
\usepackage{booktabs} % for professional tables
\usepackage{hyperref}
\usepackage{cleveref}
\usepackage{siunitx}
\usepackage{amsmath,amsfonts,bm}

\def\eqref#1{equation~\ref{#1}}
\def\1{\bm{1}}

\DeclareMathAlphabet{\mathsfit}{\encodingdefault}{\sfdefault}{m}{sl}
\SetMathAlphabet{\mathsfit}{bold}{\encodingdefault}{\sfdefault}{bx}{n}

\theoremstyle{plain}

\theoremstyle{definition}

\theoremstyle{remark}

\usepackage{hyperref}

\title{Connections between Schedule-Free Optimizers, AdEMAMix, and Accelerated SGD Variants}

\author{
Depen Morwani \thanks{Equal contribution. Correspondence to \texttt{dmorwani@g.harvard.edu}.} \\
  Kempner Institute, Harvard University \\
  \texttt{dmorwani@g.harvard.edu}
  \And
  Nikhil Vyas $ ^*$ \\
  Department of Computer Science, Harvard University \\
  \texttt{vyasnikhil96@gmail.com}
  \And
  Hanlin Zhang \\
  Department of Computer Science, Harvard University \\
  \texttt{hanlinzhang@g.harvard.edu}
  \And
  Sham Kakade \\
  Kempner Institute, Harvard University \\
  \texttt{sham@seas.harvard.edu}
}

\iclrfinalcopy % Uncomment when ready for camera submission
\begin{document}

\maketitle

\begin{abstract}
Recent advancements in deep learning optimization have introduced new algorithms, such as Schedule-Free optimizers, AdEMAMix, MARS and Lion which modify traditional momentum mechanisms. In a separate line of work, theoretical acceleration of stochastic gradient descent (SGD) in noise-dominated regime has been achieved by decoupling the momentum coefficient from the current gradient's weight. In this paper, we establish explicit connections between these two lines of work. We substantiate our theoretical findings with preliminary experiments on a 150m language modeling task. We find that AdEMAMix, which most closely resembles accelerated versions of stochastic gradient descent, exhibits superior performance. Building on these insights, we introduce a modification to AdEMAMix, termed Simplified-AdEMAMix, which maintains the same performance as AdEMAMix across both large and small batch-size settings while eliminating the need for two different momentum terms. The code for Simplified-AdEMAMix is available on the repository: \href{https://github.com/DepenM/Simplified-AdEMAMix/}{https://github.com/DepenM/Simplified-AdEMAMix/}.

\end{abstract}

\section{Introduction}
Recently, numerous optimization algorithms have been introduced for deep learning such as Lion~\citep{lion}, ScheduleFreeSGD/AdamW~\citep{schedulefree}, and AdEMAMix~\citep{ademamix}. While these optimizers have been proposed with distinct motivations, they share a common characteristic: each modifies the momentum scheme employed in optimization.

A separate body of theoretical research has focused on accelerating gradient descent in noisy environments. Although classical momentum methods, such as heavy-ball or Nesterov momentum, are sufficient to accelerate deterministic gradient descent (particularly for quadratic functions), they do not accelerate SGD~\citep{jain18accelerate, Liu2020Accelerating}. This limitation has led to the development of alternative momentum schemes aimed at achieving acceleration in the presence of noise\citep{jain18accelerate, vaswaniconvergence, Liu2020Accelerating, agnes}. Notably, all proposed accelerated SGD methods can be interpreted as decoupling the momentum coefficient from the weight assigned to the current gradient in the optimizer update.

Our primary contribution is to establish a direct connection between the ideas developed in these two research directions. Specifically, we demonstrate that Schedule-Free SGD is mathematically equivalent to performing accelerated SGD followed by weight averaging. Furthermore, optimizers such as Lion, Schedule-Free AdamW, and AdEMAMix can be understood as combining preconditioning techniques with accelerated SGD approaches. While certain aspects of these connections have been noted in prior literature~\citep{defazio2021momentumprimalaveragingtheoretical}, to the best of our knowledge, the relationship between these recently proposed optimizers and accelerated SGD has not been formally established before. 

To validate our theoretical findings, we conduct experiments using a 150m decoder-only transformer model, trained on 15b tokens with a small batch size of 32k tokens, ensuring that the training process operates in a noise-dominated regime. As predicted by our theoretical insights, the performance of Schedule-Free AdamW closely aligns with that of accelerated SGD-based AdamW~(\Cref{alg:acc_adam}). Additionally, we observe that accelerated methods offer slightly improved performance at small batch sizes. However, we also demonstrate that these performance benefits diminish at sufficiently large batch sizes, which is consistent with the theoretical connections to accelerated SGD.

Our main contributions are stated below:

\begin{enumerate} 
\item We establish precise theoretical connections between accelerated SGD and recently proposed optimizers, such as Schedule-Free SGD and AdEMAMix. 
\item We provide empirical validation through experiments on a 150-million-parameter decoder-only transformer, comparing AdamW, Schedule-Free AdamW, AdEMAMix, and MARS. Our findings indicate that AdEMAMix, which most closely aligns with accelerated SGD variants, demonstrates superior performance among these methods. 
\item As anticipated from its equivalence to accelerated SGD, the performance advantages of these methods diminish at large batch sizes relative to Adam. Notably, we show that Adam with momentum scheduling can match the performance of AdEMAMix. 
\item At high batch sizes, we observe that Schedule-Free AdamW performs significantly worse than AdamW with cosine decay, which we attribute to the intrinsic coupling of momentum and weight averaging coefficients in Schedule-Free optimizers. 
\item We introduce a modification to AdEMAMix, termed Simplified-AdEMAMix, which preserves the performance of AdEMAMix across both large and small batch size regimes, while eliminating the need for two distinct momentum terms. 
\end{enumerate}

\section{Related Work}
We review the existing literature on accelerated SGD variants and optimization algorithms that are directly relevant to our work.

\citet{jain18accelerate} introduced an accelerated SGD variant that demonstrated improved convergence rates for the least-squares problem. \citet{kidambi2018on} further simplified the update rule for this variant and formally established that momentum does not provide acceleration in this specific case. Subsequent works~\citep{Liu2020Accelerating, vaswaniconvergence, agnes} extended these results to general convex and strongly convex functions under various theoretical assumptions.

Over the years, several optimizers have been proposed that exhibit similarities to the accelerated SGD variants described above. \citet{lucas2018aggregated} introduced a method that incorporates a weighted sum of multiple momentum terms, each with distinct coefficients, to compute the final update. \citet{ma2018quasihyperbolic} developed an optimizer explicitly inspired by the theoretical framework established in \citet{jain18accelerate}. More recently, \citet{lion} proposed an optimizer discovered via a genetic search algorithm, which, similar to previous accelerated SGD variants, assigns different weights to the gradient and the momentum coefficient in the update step. Additionally, \citet{ademamix} introduced a method that blends two distinct momentum scales in the final update.

\section{Background}
\subsection{Momentum}
Momentum is a well-established technique for accelerating the convergence of gradient descent in deterministic settings. The momentum update for weights \( w_t \), with a momentum coefficient \( \beta \), is given by:

\[
m_t = \beta m_{t-1} + \nabla f(w_t); \quad w_t = w_{t-1} - \eta m_t
\]

\subsection{Weight Averaging}
Weight averaging is a widely used technique in stochastic optimization to reduce noise in the iterates. Instead of returning the final iterate \( w_T \), a weighted average \( \bar{w}_T \) of the iterates is computed, where the weights are denoted by \( \gamma_t \):

\[
\bar{w}_T = (1-\gamma_T)\bar{w}_{T-1} + \gamma_T w_T
\]

All instances of weight averaging in this paper utilize coefficients \( \gamma_t \) of the form:

\[
\gamma_t \approx 1 - \frac{1}{\delta t}
\]

for some constant \( 0 \leq \delta \leq 1 \).

\subsection{Accelerated SGD}
In this section, we provide a generalized framework encompassing many accelerated SGD methods:

\begin{equation} \label{eq:acc_sgd_form}
      m_{t} = \beta_{\text{a}, t}m_{t-1} + g_t, \quad w_{t+1} = w_t-\eta_{\text{a}, t} m_{t} - \alpha_{\text{a}, t} g_t
\end{equation}

where \( \beta_{\text{a}, t}, \alpha_{\text{a}, t}, \eta_{\text{a}, t} \) are (possibly time-dependent) scalar coefficients, and \( g_t \) represents the stochastic gradient evaluated at \( w_t \). We use the subscript ‘a’ to indicate coefficients that adhere to this specific accelerated SGD formulation.

We first note that setting \( \alpha_{\text{a}, t} = 0 \) recovers standard SGD with momentum. Additionally, as observed in prior work, many accelerated SGD algorithms proposed in the literature—such as those introduced by \citet{jain18accelerate, vaswaniconvergence, Liu2020Accelerating, agnes}—fall directly within this framework. A precise demonstration of this equivalence is provided in Appendix \ref{app:equivalence}.

\section{Connections between Existing Optimizers and Accelerated SGD}
In this section, we theoretically establish precise connections between existing optimizers, such as Schedule-Free optimizers and AdEMAMix, and accelerated SGD. Based on these insights, we propose a simplified variant of AdEMAMix that utilizes a single momentum term while maintaining performance comparable to AdEMAMix across both small and large batch size regimes.

\subsection{Schedule-Free SGD}
Schedule-Free SGD~\citep{schedulefree} is a recently introduced constant learning rate optimizer designed to eliminate the need for scheduling. Following the notation used in \citet{schedulefree}, the update equations are given by:

\begin{align*}
y_t &= (1-\beta)z_t + \beta x_t \\
z_{t+1} &= z_t - \gamma g(y_t) \\
x_{t+1} &= (1-c_{t+1})x_t + c_{t+1} z_{t+1}
\end{align*}

Here, \( y_t \) represents the current model weights (where the gradient is evaluated), while \( x_t \) denotes the weights used for evaluation.

We first express the update in terms of \( y_t \) and \( m_t \), where we define:

\begin{align}
	m_{t+1} &= \frac{x_t - z_{t+1}}{\gamma}.
\end{align}

Further simplifying \( m_{t+1} \), we obtain:

\begin{align}
	m_{t} &= \frac{x_t - z_{t+1}}{\gamma} \\
	&=  \frac{x_{t} + \gamma g_t - z_{t}}{\gamma} \\
	&=  \frac{x_{t} - z_t +  \gamma g_t}{\gamma} \\
	&=  \frac{(1-c_{t})(x_{t-1}-z_{t}) +  \gamma g_t}{\gamma} \\
	&= (1 - c_{t})m_{t-1} + g_t.
\end{align}

Thus, \( m_t \) follows the momentum update in \Cref{eq:acc_sgd_form} with \( \beta_{\text{a},t} = 1-c_t \). Given \( m_t \), we now examine the update for \( y_t \):

\begin{align}
y_{t+1} &= (1 - \beta)z_{t+1} + \beta x_{t+1} \\
        &= (1 - \beta)\left(z_t - \gamma g_t\right) + \beta \left((1 - c_{t+1})x_t + c_{t+1} z_{t+1}\right) \\
        &= (1 - \beta)z_t + \beta x_t - (1 - \beta)\gamma g_t + \beta c_{t+1} (z_{t+1} - x_t) \\
        &= y_t - \gamma[\beta c_{t+1} m_{t} + (1 - \beta) g_t].
\end{align}

Thus, \( y_t \) follows the weight update in \Cref{eq:acc_sgd_form} with \( \eta_{\text{a}, t} = \gamma \beta c_{t+1} \) and \( \alpha_{\text{a}, t} = \gamma(1-\beta) \), where \( w_t = y_t \). Consequently, \( y_t \) in Schedule-Free SGD precisely follows the accelerated SGD framework. However, \( x_t \) is used for evaluation in Schedule-Free SGD. We now analyze the dynamics of \( x_t \):

\begin{align*}
	x_{t+1} &= (1-c_{t+1})x_t + c_{t+1} z_{t+1}\\
	x_{t+1} &= (1 - c_{t+1})x_t + c_2 \left(\frac{y_{t+1} - \beta x_{t+1}}{1 - \beta}\right) \\
	x_{t+1} \left(1 - \beta + c_{t+1} \beta\right) &= (1 - c_{t+1})(1 - \beta)x_t + c_{t+1} y_{t+1} \\
	x_{t+1} &= \frac{(1 - c_{t+1})(1 - \beta)x_t + c_{t+1} y_{t+1}}{(1 - c_{t+1})(1 - \beta)+c_{t+1}}.
\end{align*}

Thus, \( x_t \) is a weighted average of \( y_t \). Recursively expanding \( x_t \) confirms that it is an exponential average of \( y_t \) when \( c_t \) is a constant. This establishes that Schedule-Free SGD can be understood as accelerated SGD followed by weight averaging.

The benefits of Schedule-Free SGD can be attributed to two key components:
\begin{enumerate}
    \item Improved performance compared to standard SGD with momentum, due to its equivalence to accelerated SGD.
    \item The ability to use a constant learning rate without scheduling, enabled by weight averaging (specifically, tailed weight averaging; see \Cref{sec:def_beta_schedfree}).
\end{enumerate}

We note two advantages unique to Schedule-Free SGD/Adam:
\begin{itemize}
    \item It does not require additional memory for weight averaging.
    \item It eliminates the need for an explicit weight averaging coefficient as a hyperparameter.
\end{itemize}

However, in \Cref{sec:largebsz}, we demonstrate that this coupling of momentum and weight averaging coefficients does not scale well for large batch sizes.

\subsubsection{Case: \( \beta = 0.0 \)}
As noted in~\citep{schedulefree}, when \( \beta = 0 \), Schedule-Free SGD reduces to standard SGD with weight averaging. Since \( c_t = 1/t \), it applies weight averaging from the beginning.

\subsubsection{Case: \( \beta = 1.0 \)}
As noted in~\citep{schedulefree}, when \( \beta = 1 \), Schedule-Free SGD reduces to standard momentum SGD, with the momentum coefficient \( \beta_{\text{a}, t} \) scaling as \( 1 - 1/t \). 

\subsubsection{Case: \( \beta = 0.9 \)} \label{sec:def_beta_schedfree}
For \( \beta = 0.9 \), the default setting in Schedule-Free SGD:
\begin{itemize}
	\item As \( c_{t} \) scales as \( 1/t \), momentum grows as \( 1 - 1/t \).
	\item The ratio of the weight assigned to the current gradient versus momentum is fixed at \( (1-\beta)/(\beta c_{t+1})  \approx 0.11 \).
	\item Weight averaging is applied approximately over the most recent \( 10\% \) of the iterates.
\end{itemize}

\subsection{Lion}
The update rule for Lion~\citep{lion} is given by:

\[
m'_t = \beta_1 m_{t-1} + (1-\beta_1) g_t
\]

\[
\theta_t = \theta_{t-1} - \eta \text{sign}(m'_t)
\]

\[
m_t = \beta_2 m_{t-1} + (1-\beta_2) g_t.
\]

Lion~\citep{lion} can be directly interpreted as an accelerated SGD method followed by a coordinate-wise sign operation.

\subsection{MARS}
In this section, we demonstrate that the practical version of the recently proposed optimizer MARS~\citep{mars2024}, referred to as MARS-Approx, follows the accelerated SGD framework, supplemented by a preconditioning step. The update equations (ignoring bias correction and clipping) are given by:

\begin{align*}
c_t &= g_t + \gamma \frac{\beta_1}{1- \beta_1}[g_t - g_{t-1}] \\
m_t &= \beta_1m_{t-1} + (1-\beta_1)c_t \\
v_t &= \beta_2 v_{t-1} + (1-\beta_2) c_t^2 \\
x_{t+1} &= x_t - \eta \frac{m_t}{\sqrt{v_t}+\epsilon}
\end{align*}

where \( m_t \) and \( v_t \) represent the first- and second-order momentum terms, respectively, and \( x_t \) denotes the model parameters. Rewriting the update using \( \hat{m}_t = m_t - \gamma g_t \), we obtain:

\begin{align*}
    c_t &= g_t + \gamma \frac{\beta_1}{1- \beta_1}[g_t - g_{t-1}] \\
    \hat{m}_t &= \beta_1 \hat{m}_{t-1} + (1-\beta_1)(1-\gamma)g_t \\
    v_t &= \beta_2 v_{t-1} + (1-\beta_2) c_t^2 \\
    x_{t+1} &= x_t - \eta \frac{\hat{m}_t + \gamma g_t}{\sqrt{v_t}+\epsilon}
\end{align*}

This formulation illustrates that the momentum update follows the general accelerated SGD framework. However, it is important to note that MARS employs a distinct preconditioning approach compared to AdamW. We further analyze its empirical performance in \Cref{sec:exp}.

\subsection{AdEMAMix}
The recently proposed optimizer AdEMAMix~\citep{ademamix} shares structural similarities with accelerated SGD-based AdamW. However, instead of using a linear combination of the current gradient and the momentum term as in accelerated SGD, AdEMAMix maintains two distinct momentum terms with different coefficients and computes their linear combination. The algorithm is formally stated in \Cref{alg:ademamix}.

To simplify our analysis, we consider a variant of AdEMAMix with \( \beta_1 = 0 \). As demonstrated in \citet{ademamix}, this simplified version achieves performance nearly equivalent to the full version for small batch sizes. Our experiments in \Cref{sec:exp} corroborate this finding. With \( \beta_1 = 0 \), AdEMAMix aligns with the general accelerated SGD framework (\Cref{eq:acc_sgd_form}). Furthermore, we show that the prescribed schedules for \( \beta_3 \) (momentum coefficient) and \( \alpha \) (which controls the relative weight assigned to the current gradient) in AdEMAMix closely match theoretical schedules proposed for accelerated SGD~\citep{agnes}.

In smooth convex optimization, achieving acceleration in stochastic settings requires a momentum scheme of the form:

\[
\beta_{a,t} = 1 - \frac{k}{t}
\]

for some constant \( k > 0 \), as established by \citet{agnes}. The AdEMAMix optimizer approximately follows this scheme by scaling up \( \beta_3 \) accordingly.

Additionally, note that in accelerated SGD schemes, momentum is maintained in the standard form:

\[
m_t = \beta_{a,t} m_{t-1} + g_t
\]

whereas in \Cref{alg:ademamix}, the momentum update follows:

\[
m_2^{(t)} \gets \beta_3^{(t)} m_2^{(t-1)} + (1 - \beta_3^{(t)}) g^{(t)}.
\]

For \( \beta_{a,t} \) scaling as \( 1 - 1/t \), the accumulated contribution of past gradients in \( m_t \) in accelerated SGD grows proportionally to \( t \). Similarly, the coefficient \( \alpha \) in AdEMAMix also scales proportionally to \( t \). Due to these similarities, AdEMAMix demonstrates improved empirical performance relative to other optimizers, as observed in \Cref{fig:best_val}.

For large batch sizes, however, AdEMAMix exhibits a performance decline when using \( \beta_1 = 0.0 \), as reported in \citet{ademamix}. \citet{agnes} suggests that for large batch sizes, the weight assigned to the current gradient in the update must decrease. In contrast, AdEMAMix maintains a fixed weight of 1 on the current gradient, which likely contributes to its diminished performance at large batch sizes.

In the following section, we introduce a simplified variant of AdEMAMix that incorporates a weight on the current gradient, removes the need for scheduling \( \alpha \) and maintains only a single momentum term. We empirically validate that this simplified version performs comparably to AdEMAMix across both small and large batch setups.

\begin{algorithm}
\caption{Single step of AdEMAMix optimizer.}\label{alg:ademamix}
\begin{algorithmic}[1]
\STATE \textbf{Input:} Data distribution $\mathcal{D}$. Initial model parameters $\theta^{(0)}$. Number of iterations $T$. Learning rate $\eta$. $\epsilon$ a small constant. AdamW parameters: $\beta_1$, $\beta_2$. AdEMAMix parameters $\beta_3$, $\alpha$. Warmup parameter $T_{\alpha, \beta_3}$, note that we usually set it to $T$. $\beta_{\text{start}}$ is usually set to $\beta_1$.
    \STATE Optional: use schedulers $\eta^{(t)}$, $\beta_3^{(t)} \gets f_{\beta_3}(t, \beta_3, \beta_{\text{start}}, T_{\alpha, \beta_3})$ and $\alpha^{(t)} \gets f_{\alpha}(t, \alpha, T_{\alpha, \beta_3})$
    \STATE Sample batch: $x \sim \mathcal{D}$
    \STATE Compute gradient: $g^{(t)} \gets \nabla_\theta \mathcal{L}_{\theta^{(t-1)}}(x)$
    \STATE Update the fast EMA $m_1$: $m_1^{(t)} \gets \beta_1 m_1^{(t-1)} + (1 - \beta_1) g^{(t)}$
    \STATE Update the slow EMA $m_2$: $m_2^{(t)} \gets \beta_3^{(t)} m_2^{(t-1)} + (1 - \beta_3^{(t)}) g^{(t)}$
    \STATE Update the second moment estimate: $\nu^{(t)} \gets \beta_2 \nu^{(t-1)} + (1 - \beta_2) \left(g^{(t)}\right)^2$
    \STATE Update parameters: $\theta^{(t)} \gets \theta^{(t-1)} - \eta^{(t)} \left(\frac{\hat{m}_1^{(t)} + \alpha^{(t)} m_2^{(t)}}{\sqrt{\hat{\nu}^{(t)}} + \epsilon}\right)$
\end{algorithmic}
\end{algorithm}

\begin{algorithm}
\caption{Single step of Simplified AdEMAMix optimizer.}\label{alg:simp_ademamix}
\begin{algorithmic}[1]
\STATE \textbf{Input:} Data distribution $\mathcal{D}$. Initial model parameters $\theta^{(0)}$. Number of iterations $T$. Learning rate $\eta$. $\epsilon$ a small constant. AdamW parameters: $\beta_1$, $\beta_2$. AdEMAMix parameters $\alpha, \beta_{\text{start}}$. Warmup parameter $T_{\beta_1}$, note that we usually set it to $T$.
    \STATE Optional: use schedulers $\eta^{(t)}$, $\beta_1^{(t)} \gets f_{\beta_1}(t, \beta_1, \beta_{\text{start}}, T_{\beta_1})$
    \STATE Sample batch: $x \sim \mathcal{D}$
    \STATE Compute gradient: $g^{(t)} \gets \nabla_\theta \mathcal{L}_{\theta^{(t-1)}}(x)$
    \STATE Update the EMA $m_1$: $m_1^{(t)} \gets \beta_1 m_1^{(t-1)} + g^{(t)}$
    \STATE Update the second moment estimate: $\nu^{(t)} \gets \beta_2 \nu^{(t-1)} + (1 - \beta_2) \left(g^{(t)}\right)^2$
    \STATE Update parameters: $\theta^{(t)} \gets \theta^{(t-1)} - \eta^{(t)} \left(\frac{m_1^{(t)} + \alpha g^{(t)}}{\sqrt{\hat{\nu}^{(t)}} + \epsilon}\right)$
\end{algorithmic}
\end{algorithm}

\subsection{Simplified AdEMAMix}
Building on the insights discussed above, we propose a simplified optimizer that eliminates the need for maintaining two separate momentum terms and removes the requirement for scheduling \( \alpha \). The optimizer is formally presented in \Cref{alg:simp_ademamix}, where we employ theory-style momentum (instead of the exponential moving average (EMA) style). In the final update, we assign a fixed weight \( \alpha \) to the gradient. We note that setting \( \alpha = 0 \) recovers the standard Adam optimizer (subject to appropriate transformations of \( \eta \) and \( \beta_1 \)). In \Cref{sec:exp}, we demonstrate that this simplified variant matches the performance of AdEMAMix across both small and large batch sizes.

\section{Experiments}
\label{sec:exp}

\begin{figure*}[!h]
    \centering
    \includegraphics[width=0.7\linewidth]{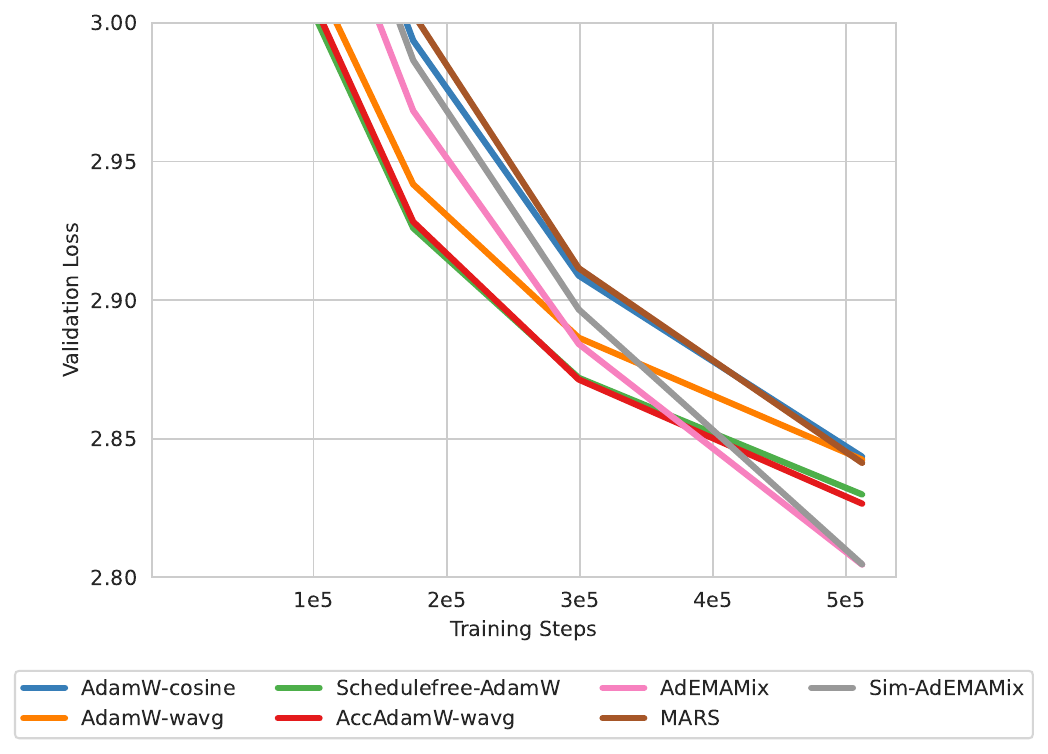}
    \caption{Comparison of the best runs of various optimizers as stated in Section \ref{sec:exp} for language modeling task on a decoder-only 150m transformer model. We find that AdEMAMix and simplified-AdEMAMix perform the best, owing to their precise similarity to accelerated SGD variants.}
    \label{fig:best_val}
\end{figure*}

\begin{algorithm}[t]
	
	\begin{algorithmic}[1]
		\STATE Sample batch $B_t$.
		\STATE $g \gets -\nabla_w \phi_{B_t}(w_t)$
		\STATE $v \gets \beta_2 v + (1-\beta_2) (g \odot g)$
		\STATE $N \gets \frac{\beta_3 m + (1-\beta_3)g}{\sqrt{\hat{v}} + \epsilon}$ 
		\STATE $w \gets w - \eta N$
            \STATE $m \gets \beta_1 m + (1-\beta_1) g$
            \STATE $c = \max(1-1/t, 1-1/(\delta t))$
            \STATE $w_\text{avg} \gets c  w_\text{avg}+(1-c)w$
	\end{algorithmic}
	\caption{Single step of accelerated SGD based Adam with weight averaging. For simplicity we ignore the initialization, other boundary effects such as bias correction, and weight decay. Hyperparameters: Learning rate $\eta$, $\text{betas} = (\beta_1, \beta_2, \beta_3)$, weight averaging coefficient $\delta$, and epsilon $\epsilon$.}
	\label{alg:acc_adam}
\end{algorithm}

In this section, we present experiments conducted on a 150-million-parameter decoder-only transformer model for a language modeling task using the C4 dataset. The model is trained with a sequence length of 1024 and a batch size of 32, over 15 billion tokens (\(\approx 5 \times\) Chinchilla), ensuring that the training operates in a noise-dominated regime. 

We compare the following optimization algorithms:

\begin{enumerate}
    \item Standard AdamW with cosine decay
    \item Standard AdamW with weight averaging
    \item Schedule-Free AdamW
    \item Accelerated AdamW with weight averaging (\Cref{alg:acc_adam})
    \item MARS
    \item AdEMAMix
    \item Simplified-AdEMAMix
\end{enumerate}

Details of hyperparameter sweeps for these algorithms are provided in \Cref{app:hyperparameters}.

As illustrated in \Cref{fig:best_val}, Schedule-Free AdamW and Accelerated AdamW with tailed weight averaging perform comparably, supporting our theoretical claims. Furthermore, both outperform AdamW with cosine decay and AdamW with tailed weight averaging. Moreover, AdEMAMix and Simplified-AdEMAMix outperform all methods, which we hypothesize is due to their alignment with accelerated SGD variants.

\subsection{Large Batch Size Experiments}
\label{sec:largebsz}

\begin{figure*}[t]
    \centering
    \includegraphics[width=0.6\linewidth]{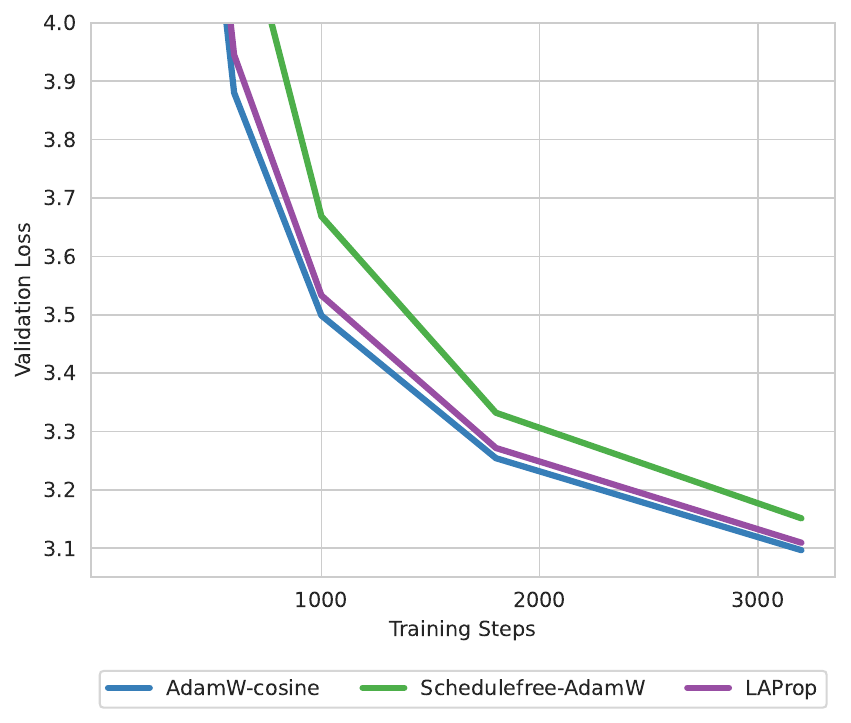}
    \caption{Comparison of the best runs of AdamW with cosine decay, schedule free AdamW and LAProp at higher batch size. Experimental details can be found in Section \ref{sec:largebsz}}
    \label{fig:adamw_sf-adamw}
\end{figure*}

\begin{figure*}[t]
    \centering
    \includegraphics[width=0.6\linewidth]{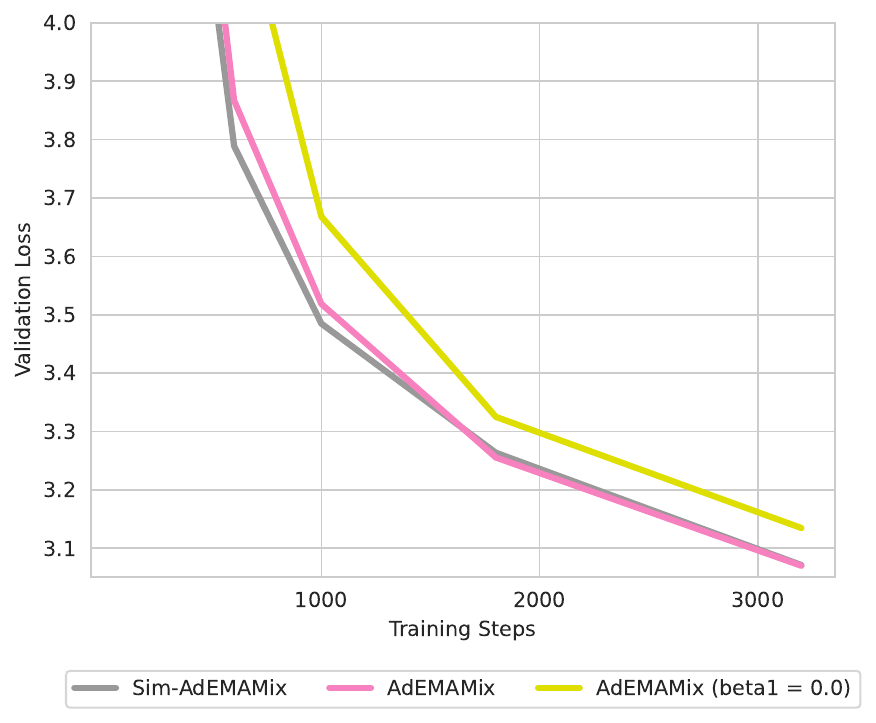}
    \caption{Comparison of the best runs of AdEMAMix (with and without $\beta_1 = 0.0$) and our variant of simplified AdEMAMix for higher batch size experiments. Experimental details can be found in Section \ref{sec:largebsz}}
    \label{fig:adema_large_batch}
\end{figure*}

While the previous experiments focused on the small batch size regime (i.e., training with noisy gradients), we now conduct experiments in the large batch size regime to assess whether these algorithms generalize effectively. In this setup, we train models with a batch size of 1 million tokens over 3 billion tokens (\(\approx\) Chinchilla scale).

\paragraph{Schedule-Free AdamW:} As shown in \Cref{fig:adamw_sf-adamw}, Schedule-Free AdamW performs significantly worse compared to AdamW. We attribute this performance gap to the coupling between weight averaging and momentum coefficients. At higher batch sizes, the optimal momentum value is significantly lower than \( 1 - 1/t \). Although one could use a scaling factor \( \approx 1 - r/t \) for some \( r \geq 1 \), a higher \( r \) reduces the effective weight averaging window.

Another key distinction between AdamW and Schedule-Free AdamW is the order in which momentum and preconditioning are applied. AdamW applies momentum before preconditioning, whereas Schedule-Free AdamW applies preconditioning before momentum, making it algorithmically similar to LAProp~\citep{laprop2021}. However, as shown in \Cref{fig:adamw_sf-adamw}, the performance of AdamW is comparable to that of LAProp, suggesting that this difference is not the primary cause of the performance gap.

\paragraph{AdEMAMix:} For large batch sizes, as previously observed in \citet{ademamix}, Figure \ref{fig:adema_large_batch} shows that setting \( \beta_1 = 0.0 \) in AdEMAMix results in a significant performance drop compared to using two separate momentum terms. This degradation occurs because AdEMAMix assigns a fixed weight of 1 to the current gradient, whereas theoretical accelerated SGD variants~\citep{agnes} require a diminishing weight on the current gradient as batch size increases. 

Additionally, as depicted in Figure \ref{fig:adema_large_batch}, our proposed variant, Simplified-AdEMAMix, achieves performance equivalent to AdEMAMix while eliminating the need for two separate momentum terms. Notably, we achieve this performance at \( \alpha = 0.0 \), meaning Simplified-AdEMAMix reduces to standard Adam with momentum scheduling.

\section{Conclusion}
In this work, we establish explicit connections between accelerated SGD variants and several recently proposed optimizers, including Schedule-Free optimizers, AdEMAMix, MARS, and Lion. We also present empirical evidence demonstrating that AdEMAMix, which aligns most closely with theoretical accelerated SGD variants, achieves superior performance in small batch size training.

Building on this connection, we introduce Simplified-AdEMAMix, which removes the need for maintaining two separate momentum buffers. We empirically show that Simplified-AdEMAMix matches the performance of AdEMAMix across both small and large batch sizes while eliminating the additional memory overhead associated with AdEMAMix.

\section*{Acknowledgments}

SK, HZ, and DM acknowledge support from the Office of Naval Research under award N0001422-1-2377 and the National Science Foundation Grant under award \#IIS 2229881. This work has been made possible in part by a gift from the Chan Zuckerberg Initiative Foundation to establish the
Kempner Institute for the Study of Natural and Artificial Intelligence. NV and DM are supported
by a Simons Investigator Fellowship, NSF grant DMS-2134157, DARPA grant W911NF2010021,and
DOE grant DE-SC0022199.

%we showed the equivalence between the recently proposed schedule-free SGD optimizer and accelerated SGD with weight averaging. We also empirically demonstrated that accelerated SGD demonstrates improved performance as compared to SGD with momentum for training a 150m decoder-only transformer model on language modeling task.

% {\color{blue}

% \begin{itemize}
%     \item ScheduleFreeAdam does LA prop style does it hurt at large batch sizes.  Find which variant of AGNES+ADam works best at both small and large batch sizes. 
% \end{itemize}

% }

\bibliography{sample}
\bibliographystyle{icml2025}

\newpage
\appendix
\onecolumn

\section{Hyperparameters} \label{app:hyperparameters}
Below are the hyperparameters for the small batch experiments.
\begin{enumerate}
    \item AdamW with cosine decay - 51.2k warmup -  learning rate in [3.16e-4, 1e-3, 3.16e-3], $\beta_1$ in [0.9, 0.95], $\beta_2$ in [0.99, 0.999, 0.99968, 0.9999]. The optimal values of $\beta_1$ and $\beta_2$ were $.9$ and $.999$ respectively matching the default values. We note that for larger batch sizes it is common to use $\beta_2 = .95$, the benfit of higher $\beta_2$ at smaller batch sizes has also been observed by~\citet{porian}. 
    \item AdamW with cosine decay - 10k warmup -  learning rate in  [3.16e-4, 1e-3, 3.16e-3], $\beta_1 = 0.9$, $\beta_2 = 0.999$ i.e. we fix $\beta_1, \beta_2$ to be the optimal values from the previous sweep. This performed worse that warmup of 51.2k steps. 
    \item AdamW constant fraction weight averaging: - learning rate in [3.16e-4, 1e-3, 3.16e-3], $\beta_1 = 0.9$, $\beta_2$ in
[0.99, 0.997, 0.999, 0.9997], $\delta$ in [0.05, 0.1, 0.2].
    \item AdamW with cosine decay and weight averaging - learning rate in [3.16e-4, 1e-3, 3.16e-3], $\beta_1 = 0.9$, $\beta_2 = 0.999$, $\delta$ in [0.025, 0.05, 0.1].
    \item Accelerated SGD based AdamW with cosine decay - learning rate in [3.16e-4, 1e-3, 3.16e-3], $\beta_1$ in [0.999, 0.99968, 0.9999], $\beta_2$ in  [0.99, 0.9968, 0.999],  $\beta_3 = 0.9$
    \item Accelerated SGD based AdamW with constant learning rate and weight averaging - learning rate in [3.16e-4, 1e-3, 3.16e-3], $\beta_1$ in [0.99684, 0.999], $\beta_2$ in [0.999], $\beta_3 = 0.9$, $\delta$ in [0.05, 0.1]
    \item Accelerated SGD based AdamW with cosine decay and weight average - learning rate in [3.16e-4, 1e-3, 3.16e-3], $\beta_1$ in [0.99684, 0.999], $\beta_2 = 0.999$, $\delta$ in [0.05, 0.1], $\beta_3 = 0.9$
    \item Schedulefree AdamW with constant learning rate - learning rate in [3.16e-4, 1e-3, 3.16e-3, 1e-2], $\beta_1$ in [0.8, 0.9, 0.95], $\beta_2 = 0.999$
    \item Schedulefree AdamW with cosine decay - [3.16e-4, 1e-3, 3.16e-3, 1e-2], $\beta_1$ in [0.8, 0.9, 0.95], $\beta_2 = 0.999$
    \item MARS - [3.16e-4, 1e-3, 3.16e-3, 1e-2], $\beta_1$ in [0.9, 0.95 0.99], $\beta_2$ in [0.99, 0.999], $\gamma$ in [0.0, 0.01, 0.02, 0.03, 0.04, 0.05], precondition 1d was set to True.
    \item AdEMAMix - [3.16e-4, 1e-3, 3.16e-3], $\beta_1$ in [0.0, 0.9], $\beta_2 = 0.999$, $\beta_3$ in  [0.99, 0.999, 0.9999], $\alpha$ in [2,4,8,16].
    \item Sim-AdEMAMix - [1e-6, 3.16e-6, 1e-5, 3.16e-5], $\beta_1$ in [0.99, 0.999, 0.9999], $\beta_2 = 0.999$, $\alpha$ in [10, 20, 50, 100] 
\end{enumerate}

The hyperparameter sweeps for the large batch experiments are provided below:

\begin{enumerate}
    \item Schedule-Free AdamW: [1e-3, 3.16e-3, 1e-2], $\beta$ in [0.8,0.9,0.95], $\beta_2$ in [0.9,0.95], $r$ in [0.0, 5.0, 9.0, 50.0]
    \item AdamW: [1e-3, 3.16e-3, 1e-2], $\beta_1$ in [0.9,0.95], $\beta_2$ in [0.9, 0.95]
    \item LAProp: [1e-3, 3.16e-3, 1e-2], $\beta_1$ in [0.9,0.95], $\beta_2$ in [0.9, 0.95]
    \item AdEMAMix: [1e-3, 3.16e-3, 1e-2], $\beta_1$ in [0.0, 0.9], $\beta_2 = 0.95$, $\beta_3$ in [0.9, 0.95, 0.99], $\alpha$ in [2,4,8,16]
    \item Sim-AdEMAMix: [1e-4, 3.16e-4, 1e-3], $\beta_1$ in [0.9, 0.95, 0.99], $\beta_2 = 0.95$, $\alpha$ in [0.0, 0.5, 1.0]
\end{enumerate}

\section{Equivalence of previous acceleration methods} \label{app:equivalence}
The general accelerated SGD form is provided in \Cref{eq:acc_sgd_form}. In this section, we will show that all the methods in the works \citet{jain18accelerate, vaswaniconvergence, Liu2020Accelerating, agnes} fall within this form.

\subsection{AGNES}
The update for \citet{agnes} is given below:

\begin{equation*}
    x_n' = x_n + \alpha v_n \qquad x_{n+1} = x_n' - \eta g_n' \qquad v_{n+1} = \rho_n (v_n - g_n')
\end{equation*}
where $g_n'$ is stochastic gradient evaluated on $x_n'$ and the final function is evaluated on $x_n$. The above equations can be rewritten as

\begin{equation*}
    x_{n+1}' = x_n' - \eta g_n' + \alpha v_{n+1} \qquad -\frac{v_{n+1}}{\rho_n} = \rho_{n-1} \left(-\frac{v_n}{\rho_{n-1}}\right) + g_n'
\end{equation*}
Thus $x_{n+1}'$ follows update equation of the form of \Cref{eq:acc_sgd_form}.

\subsection{ASGD}
The update for \citet{jain18accelerate} is given by:

\begin{equation*}
   y_{j-1} = \alpha x_{j-1} + (1-\alpha) v_{j-1} \qquad x_j = y_{j-1} - \delta g_{j-1} \qquad z_{j-1} = \beta y_{j-1} + (1-\beta) v_{j-1} \qquad v_j = z_{j-1} - \gamma g_{j-1} 
\end{equation*}
where $g_{j-1}$ represents the stochastic gradient evaluated on $y_{j-1}$ and the function is evaluated on the tail averaged $x$.

The update equations above can be rewritten as:
\begin{equation*}
    y_j = y_{j-1} - \alpha \delta g_{j-1} - (1-\alpha)[y_{j-1} - v_j] \qquad \frac{y_{j-1} - v_j}{\gamma - (1-\beta)\alpha\delta} = (1-\beta)\alpha \frac{y_{j-2} - v_{j-1}}{\gamma - (1-\beta)\alpha\delta} + g_{j-1}
\end{equation*}
The update equations above follow the form of \Cref{eq:acc_sgd_form}.

\subsection{MaSS}
The update for \citet{Liu2020Accelerating} is given by:

\begin{equation*}
    w_{t+1} = u_t - \eta_1 g_t \qquad u_{t+1} = (1 + \gamma)w_{t+1} - \gamma w_t + \eta_2 g_t
\end{equation*}

where $g_t$ is the stochastic gradient evaluated on $u_t$ and the function is evaluated on $w_t$. These equations can be rewritten as:

\begin{equation*}
    u_{t+1} = u_t - \gamma (w_t - u_t) - [\eta_1(1+\gamma) - \eta_2] g_t \qquad \frac{w_t - u_t}{\eta_1 \gamma - \eta_2} = \gamma \frac{w_{t-1} - u_{t-1}}{\eta_1\gamma - \eta_2} + g_t
\end{equation*}
The update equations above follow the form of \Cref{eq:acc_sgd_form}.

\subsection{SGD with Nesterov Acceleration}
The update for \citet{vaswaniconvergence} is given by:

\begin{equation*}
    w_{k+1} = \zeta_k - \eta g_k \qquad \zeta_k = \alpha_k v_k + (1-\alpha_k)w_k \qquad v_{k+1} = \beta_k v_k + (1-\beta_k)\zeta_k - \gamma_k \eta g_k
\end{equation*}

These equations can be rewritten as:

\begin{equation*}
    \zeta_{k+1} = \zeta_k - \eta g_k + \alpha_{k+1}[v_{k+1} - w_{k+1}]; \qquad v_{k+1} - w_{k+1} = \beta_k (1 - \alpha_k)[v_k - w_k] - \eta (\gamma_k - 1) g_k
\end{equation*}


% You can have as much text here as you want. The main body must be at most $8$ pages long.
% For the final version, one more page can be added.
% If you want, you can use an appendix like this one.  

% The $\mathtt{\backslash onecolumn}$ command above can be kept in place if you prefer a one-column appendix, or can be removed if you prefer a two-column appendix.  Apart from this possible change, the style (font size, spacing, margins, page numbering, etc.) should be kept the same as the main body.
%%%%%%%%%%%%%%%%%%%%%%%%%%%%%%%%%%%%%%%%%%%%%%%%%%%%%%%%%%%%%%%%%%%%%%%%%%%%%%%
%%%%%%%%%%%%%%%%%%%%%%%%%%%%%%%%%%%%%%%%%%%%%%%%%%%%%%%%%%%%%%%%%%%%%%%%%%%%%%%

\end{document}